\pdfoutput=1

\documentclass[11pt]{article}

\usepackage[preprint]{acl}
\usepackage{booktabs} 
\usepackage{amsmath}
\usepackage{cleveref}
 
\usepackage{times}
\usepackage{latexsym}

\usepackage[T1]{fontenc}

\usepackage[utf8]{inputenc}

\usepackage{microtype}

\usepackage{inconsolata}

\usepackage{graphicx}

\title{PerspectiveNet: Understand Multi-View Videos with Event Matching for Traffic Safety Description and Analysis}

\author{Phu-Vinh Nguyen \\
  Uppsala University \\
  \texttt{phu-vinh.nguyen.1216@student.uu.se} \\\And
  Tan-Hanh Pham \\
  Florida Institute of Technology, USA \\
  \texttt{hanhpt.phamtan@gmail.com} \\}

\begin{document}
\maketitle
\begin{abstract}
Generating detailed descriptions for events from multiple cameras and viewpoints is challenging due to the complex and inconsistent nature of the visual data. In this paper, we present PerspectiveNet, a lightweight yet efficient model to generate long descriptions for multiple-view cameras by using a vision encoder, a small connector module to convert visual features to a fixed-size tensor, and large language models (LLM) to utilize the strong capability of LLM in natural language generation. The connector module has three primary objectives: map visual features onto LLM embedding, emphasize the critical information needed for generating descriptions, and yield the fixed-size feature matrix as the output. Furthermore, we enhance the model's ability by incorporating a second task, sequence frame detection, which enables the model to search for the relevant frame sequence. Finally, we combine the connector module, a second training task, a large language model, and a visual feature extraction model into a single model and train it for the Traffic Safety Description and Analysis task, in which the model has to generate long fine-grain descriptions for an event from multiple cameras with and viewpoints. The resulting model is lightweight and efficient for both training and inference.
\end{abstract}

\section{Introduction}
\label{sec:intro}

In recent years, there has been a variety of research about large language models, improving the capabilities of LLMs in natural language understanding, reasoning, knowledge updating, and instruction following. Moreover, LLMs can be adapted to different domains by fine-tuning different datasets, which shows their potential to assist people, enhance productivity, and handle many difficult problems. However, since LLMs are just about text and can only work with language tasks, they cannot understand different information sources such as sound and vision, which are crucial to solving many real-world problems. For those reasons, there is much research about integrating LLMs with vision models to enable LLMs with visual tasks such as LLaVA~\cite{liu2023visual}, Qwen-VL~\cite{Bai2023QwenVLAV}, InstructBLIP~\cite{dai2024instructblip}, and Flamingo~\cite{alayrac2022flamingo}, which advances the LLMs' capability in visual tasks significantly.

While problems with images and videos are the main target of many vision language models, multiple viewpoints of an event are not on the priority list. As a result, most models are not effectively designed and trained to handle such visual information. Different from video data and images, which only show objects and events from one viewpoint, this type of data can provide more information about an event by inspecting it from different positions and help vision language models understand and generate descriptions about the event more precisely. While the number of videos and the frame number of each video can be various, handling such information can be more challenging than videos and images. This is because the large and undefined number of videos results in a large and undefined number of features, which leads to the difficulty in finding the similarity of relation between many events and frames in videos. Moreover, since the number of features is large, putting all of them into the LLMs to handle might not be the best choice due to the excessive increase in size of the input and the increase in computation cost. To solve this, visual features should be encoded to smaller sizes, but still need to be informative for language models to generate long, fine-grain descriptions.

In this paper, with the awareness of those problems, we introduce PerspectiveNet, a method to connect a large and undefined number of visual features with LLMs, which can be applied to many visual language tasks. The module is constructed to take visual features of multiple videos and viewpoints as input and create the output with 3 features: small, informative, and size-consistent to provide context for LLMs to generate. Moreover, we also present a second task during the training process to increase the capability of our module in detecting important events and frames. Finally, we experiment by integrating the above methods on multiple videos and viewpoints datasets, with the main task being traffic safety description and analysis.

We summarize our contributions as follows:
\begin{itemize}
    \item Create a module to connect the visual features of multiple cameras with LLMs.
    \item Propose a second task during the training process to help the model detect important features and ignore unimportant ones.
    \item We experiment with those methods in a multiple viewpoints dataset and report our results.
\end{itemize}

\section{Related work}
\label{sec:related}

In recent years, large language models (LLMs) such as Mistral~\cite{jiang2023mistral}, LLaMA~\cite{touvron2023llama}, Bloom~\cite{le2022bloom}, Phi~\cite{gunasekar2023textbooks}, and Qwen~\cite{bai2023qwen} have shown exceptional performance across various natural language processing tasks. Their success has encouraged the development of vision-language models (VLMs) that combine visual inputs with LLMs to solve tasks like visual question answering and caption generation. These VLMs generally use a vision encoder to convert visual data into feature embeddings, which are then passed to the LLM via cross-attention or adapted as contextual tokens to generate responses.

Many well-known VLMs such as LLaVA~\cite{liu2024visual}, Qwen-VL~\cite{bai2023qwen}, and InstructBLIP~\cite{dai2024instructblip} follow a modular approach by combining pretrained vision encoders (e.g., ViT~\cite{dosovitskiy2020image}) with LLMs through lightweight projection layers or cross-attention modules. More complex architectures like Flamingo~\cite{alayrac2022flamingo}, mPLUG~\cite{li2022mplug}, and BLIP~\cite{li2022blip} integrate visual features using sophisticated mechanisms such as gated cross-modal transformers or multi-level attention layers. While these methods improve reasoning and text generation performance, they often require large memory footprints (e.g., LLaVA-34B, Qwen-VL-Chat-14B) and multi-GPU infrastructure, which limits their deployment in real-world applications or resource-constrained environments.

To bridge the gap between performance and efficiency, research has explored the design of more effective "connectors" that transform visual inputs into compact yet informative representations for LLMs. One popular method is Q-Former~\cite{li2023blip}, which learns to query and distill relevant visual information from high-dimensional image features. Other works like Flamingo~\cite{alayrac2022flamingo} employ cross-attention layers to align multi-modal sequences of different lengths. Additionally, multi-task learning objectives such as image-text matching (ITM), image-text contrastive learning (ITC), and language modeling (LM) have been incorporated to guide training and improve alignment across modalities~\cite{li2022blip}. Recent studies also show that instruction tuning (e.g., InstructGPT~\cite{ouyang2022training}, OPT-IML~\cite{iyer2022opt}) significantly enhances the zero-shot and few-shot performance of VLMs.

Although some existing methods support VLMs to understand multi-camera information and have promising results, like CityLLaVA~\cite{duan2024cityllava}, Divide\&Conquer~\cite{DBLP:conf/cvpr/XuanNNXAD22}, those models still require a strong VLM baseline to produce good results. In contrast, our work introduces PerspectiveNet, a lightweight vision-language architecture tailored for multi-camera pedestrian-centered video captioning. Without relying on prior pretraining from large-scale VLMs, our method achieves competitive performance on the WTS dataset~\cite{kong2024wtspedestriancentrictrafficvideo} while maintaining a smaller size and reduced computational cost. Our method shows significantly better performance compared to methods with a similar size, like Multi-perspective+Refinement~\cite{10678626}.

\section{Approach}

In this session, we start with a brief overview of our description generation task and the dataset on which we train and test our model. After that, we introduce the details of our solution for this problem, which includes the overall architecture of the PerceptiveNet, more details of our vision language connector, and the training method.

\subsection{Task and dataset}

This paper focuses on generating long, fine-grained video descriptions for many traffic safety scenarios and pedestrian accidents. The resulting model should be able to describe the continuous moment before the incidents and normal scenes, as well as all the details about the surrounding context, attention, location, pedestrians, and vehicles. The scene of events might be recorded from different viewpoints and positions, which requires the model to understand the similarity between each viewpoint and the relevance of many frames in an event to generate informative descriptions.

The dataset used in this paper is the WTS~\cite{kong2024wtspedestriancentrictrafficvideo}. Each sample of the dataset includes 2 long and detailed descriptions for vehicle and pedestrian, $n$ videos from different viewpoints ($n\geq 1$) about a sequence of events about traffic, incident on the street, the start time and end time of the event that the model should focus on to describe, annotated segment (pre-recognition, recognition, judgment, action, and avoidance), and bounding boxes of the instance (pedestrian or vehicle) that relates to the description.

To handle this data better, we refactored the dataset structure. After this data pre-processing step, each sample of the dataset includes $n$ videos denoted as $V=[v_0, v_1, \dots,v_{n-1}]$, the frame at second $s$ of video $v_i$ is denoted as $v_{is}$, one long description $X=[x_0,x_1,...x_{t-1}]$ with $t$ is the length of description, a target object to describe $o$ (pedestrian or vehicle), start time $m$, end time $n$, segment $s$ (pre-recognition, recognition, judgment, action, and avoidance). In general, the task can be described as follows:
\begin{align}
p(X) = \prod_{i<t}{p_\theta(x_i| x_{<i}, V, o, m, n, s)}
\end{align}

Furthermore, it is worth mentioning that our solution does not use any information from the given bounding boxes, which explains why the post-processing dataset does not contain any information about bounding boxes, and even when there are many videos about an event, they all start simultaneously. For that reason, at the same time, the frames of all cameras show the same scene from different positions, and the visual context of those frames should be relatively similar. Besides multiple-view cameras, there are data samples with only one video, which makes the task for those special samples video description generation.

\subsection{Architechture}

\begin{figure*}[htbp]
  \centering
  \includegraphics[width=0.8\linewidth]{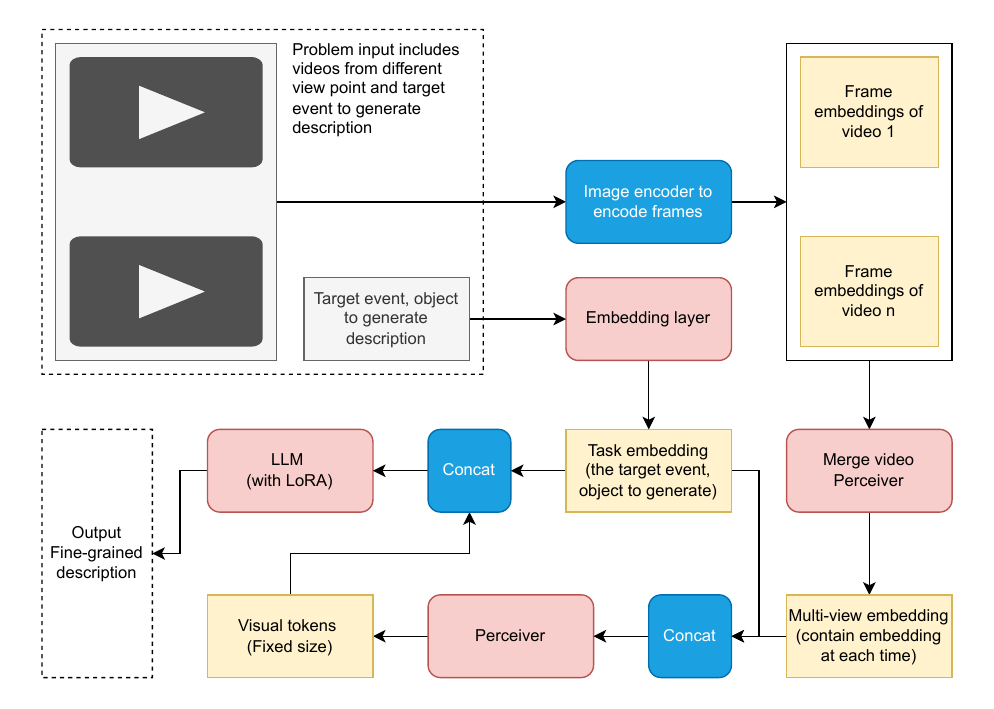}
  \caption{Architecture overview, videos are lists of frames or images. Each frame will be fed to the vision encoder to extract a list of embeddings for each video. After that, the Perceiver adapter is applied to all frames at the same moment to get the multi-view embedding. Next, it is concatenated with the task embedding before feeding it to another Perceiver layer and LLM to get the description of the event. In this image, gray color denotes input and output data, and blue denotes the frozen modules, which are not trained. Red denotes trained modules during the training phases. Finally, yellow denotes the output and input of modules.}
  \label{fig:overview_architecture}
\end{figure*}

Overall, the architecture of PerspectiveNet, as shown in \cref{fig:overview_architecture}, includes three primary components:
\begin{itemize}
    \item \textbf{Visual encoder}: to get visual information from each frame of each video, we use the pre-trained image encoder of BLIP-2\cite{li2023blip} (ViT-g/14).
    \item \textbf{Vision-Language Connector}: for the connector, we use two Perceiver module\cite{jaegle2021perceiver}. The first Perceiver is used to convert features of all frames of different videos at the same time to a single feature vector. The second Perceiver is employed to reduce the shape of the long video feature to a smaller constant shape. By doing this, we only extract important visual information and use LLM to generate descriptions. This reduces the redundancy and computation cost, ensuring the input visual context of LLM is not too long.
    \item \textbf{Large language model}: We avoid using LLMs with more than 3 billion parameters for the generating task due to the limited resources. Finally, we decided to use Phi-1.5\cite{gunasekar2023textbooks} due to its small size and strong capability in natural language generation, despite the model not having been trained on any visual information yet. Moreover, we employ LoRA\cite{hu2021lora} due to the need to fine-tune LLMs on the new task.
\end{itemize}

For each data sample, we are provided with $n$ videos, from which we extract frames from those videos at a frequency of 1 frame per second. It is important to note that even though all videos record the same event and start at the same time, not all of them have the same length; some might finish sooner than expected.

\textbf{Double Perceiver Connector}: After collecting frames from videos, we feed every single extracted frame to the visual encoder (ViT) to get visual features, the result after this step is a tensor with shape $(N_v, N_f, D)$ where $N_v$ is the number of videos, $N_f$ is the max number of frames and $D$ is the size of the visual embedding dimension, which is defined by the vision encoder. This tensor will be fed to the first Perceiver to get the new visual embedding $(1, N_f, D_1)$. Due to the context similarity between frames in all videos at the same time, this module takes the responsibility for capturing important information at that specific time. This output is then concatenated to an embedding token (which presents the object that needs to be described and the segment) to create a new tensor with shape $(1,N_f+1,D_1)$ before being fed to the second Perceiver to get a tensor of shape $(1, c, D_2)$, where $c$ is a constant. This final output is the visual summarization of the primary event in the video. Finally, we set $c=20$ in our default model.

\textbf{Task Embedding}: To generalize the model's ability to generate descriptions at each specific part of the event, we use an embedding layer to embed each task. This information is important for the model to understand what information, object, or part of the event should be focused on. In the experiment section, our dataset has five tasks and two main objects (human and vehicle), making the embedding contain ten tokens to represent all cases.

\textbf{Phi 1.5}: The new visual information is then used as a context token for Phi 1.5 to generate a description. The initial input, which provides information for LLM, includes the visual tokens, target object (pedestrian or vehicle), and segment (pre-recognition, recognition, judgment, action, or avoidance). Moreover, to make the language model adapt well to this task, LoRA\cite{hu2021lora} layers are injected into the attention mechanism of Phi 1.5. While the language model weight is frozen during training, the weights of the LoRA adapter are still set to trainable as shown in \cref{fig:llm_generation}. This step helps the language model to understand its task better after our training progress, while maintaining a strong ability in text generation.

Cross-attention layers can be added to the language model to receive visual information; however, adding this module would increase the size of the language model significantly. Furthermore, since the newly added layers are initialized with random weights, those layers will also need to be trained during the training process, which would highly increase the VRAM consumption. As a result, using visual information as extra visual context in the input prompt would be the best choice for our limited resources.

\subsection{Training Strategy}

In each data sample, we are provided with the event's start and end times that need to be described. Denote that input videos is $V=[v_0,v_1,\dots,v_{t-1}]$ where $v_i$ is all frames at time $i$, with that notation, the event from start time $m$ to end time $n$ can be written as $E=[v_m,v_{m+1},\dots,v_n]$. With two visual inputs and the requirement to generate the same description, we fed both of them to the Visual Encoder and then the adapter to collect $f_V$ and $f_E$, which are the final visual context of full video and partial events, respectively. Then, the dot product of $f_V$ and $f_E$ is calculated and compared with $I$ (Identity matrix) by cross-entropy loss.
\begin{align}
    f &= f_V \cdot f_E \\
    \mathcal{L}_M &= - \frac{1}{N} \sum_{i=1}^{N} \sum_{j=1}^{N} I_{ij} \log(f_{ij})
\end{align}
Where:
\begin{itemize}
    \item $\mathcal{L}_M$ is the loss function of matching task
    \item $N$ is the number of visual tokens
    \item $I_{ij}$ is the element in the $i$-th row and $j$-th column of the Identity matrix.
    \item $f_{ij}$ is the element in the $i$-th row and $j$-th column of $f$
\end{itemize}

Our main target for this loss function is to provide the same context (event and object to describe) but a different part of the videos (all videos and just the target event in the video) to our connector and train the connector to understand where the important event should be to generate the context. While the $f_E$ is the final feature of the correct event, the $f_V$ after training should ignore every frame, but frames can provide information about the event that needs to be described. Furthermore, the idea of using the Cross-Entropy loss in this scenario is that we want each embedding vector in the matrix should be distinguished. This means that given two random tokens in the visual embedding at different positions, their information and embeddings must be different. This ensures that the visual information will be diverse and spread across all scenes without focusing on a small part of the videos and becoming overfitted.

\begin{figure}[ht]
  \centering
  \includegraphics[width=0.9\linewidth]{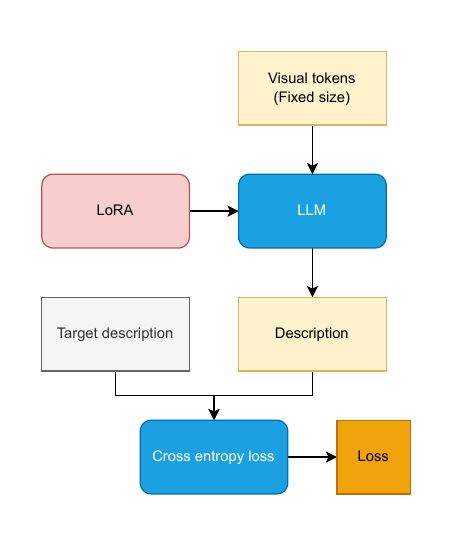}
  \caption{Training on the large language model, the model's input is only the visual tokens from previous steps. LoRa is employed to reduce the number of training parameters and to improve the model's performance on the new task. In this image, gray color denotes input and output data, and blue denotes the frozen modules, which are not trained. Red denotes trained modules during the training phases. Yellow denotes the output and input of modules. Finally, orange is the lowest value that we need to reduce.}
  \label{fig:llm_generation}
\end{figure}

\begin{figure}[ht]
  \centering
  \includegraphics[width=0.9\linewidth]{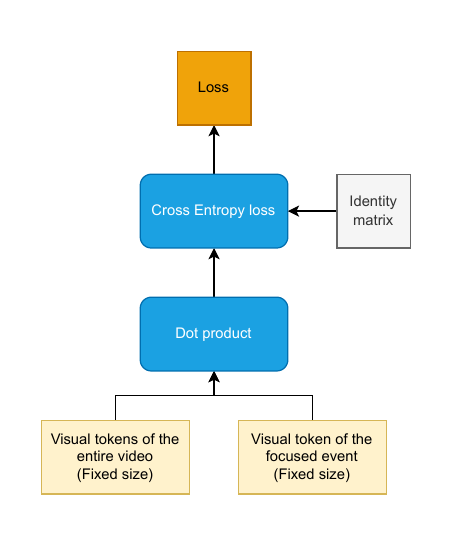}
  \caption{Second loss function on visual information to train the model to focus on important frames. The two inputs (visual tokens) are embeddings of the entire multi-view videos and of the targeted event. In this image, gray color denotes input and output data, and blue denotes the frozen modules, which are not trained. Yellow denotes the output and input of modules. Finally, orange is the loss value that we need to reduce.}
  \label{fig:second_loss}
\end{figure}

For the generation task, the prediction of visual tokens is the token with index 0. To train this model, we use LoRA to reduce the number of training parameters as shown in \cref{fig:llm_generation}. Besides using the special token for the output of visual inputs, the remaining training process, including computing loss, remained the same.
\begin{align}
\mathcal{L}_G = - \sum_{j=1}^{V} y_{j} \log(p_{j})
\end{align}
Where:
\begin{itemize}
    \item $\mathcal{L}_G$ is the loss function of generation task
    \item $V$ is the size of the vocabulary
    \item $y_j$ is the true label (one-hot encoded) for the $j$-th word
    \item $p_j$ is the predicted probability of the $j$-th word
\end{itemize}

Finally, the total loss is calculated by adding the matching loss $E_M$ and the generation loss $E_G$. By taking the backpropagation with this loss, the model can perform updates on both the generation and event matching tasks
\begin{align}
\mathcal{L}=\mathcal{L}_M+\mathcal{L}_G 
\end{align}

\section{Experiment}

\subsection{Dataset description}
\begin{figure*}[htpb]
    \centering
    \includegraphics[width=0.8\linewidth]{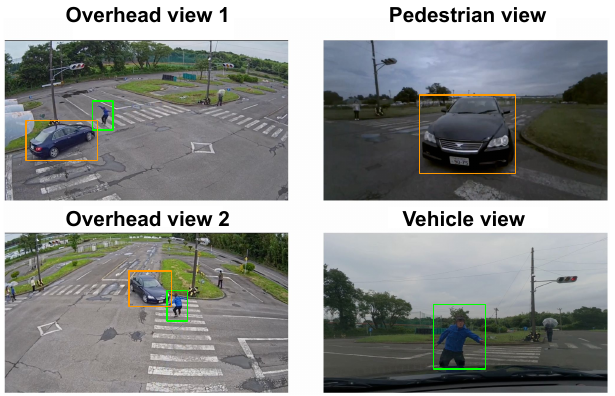}
    \caption{Example of the multi-view video in the dataset WTS at a specific time. All video in the dataset starts at the same time. However, the length of each video in a set might vary. In this sample, the correct \textbf{pedestrian behavior caption} is: The pedestrian found himself positioned directly in front of the assailant vehicle with their orientation opposite to it. There was no relative distance; they were almost touching the vehicle. His line of sight remained focused on the vehicle, closely watching its actions. The pedestrian was about to stand still after moving at a slightly higher speed, and they were decelerating. The image is collected from the original paper of WTS}
    \label{fig:data_overview}
\end{figure*}

The Woven Traffic Safety (WTS) dataset, introduced in Track 2 of the AICity Challenge 2024, is designed for fine-grained understanding of pedestrian-vehicle interactions in traffic scenarios. It contains 155 multi-view scenarios (810 videos in total), recorded at 1080p and 30 fps from overhead and vehicle-mounted cameras. Each scenario includes five temporal phases: pre-recognition, recognition, judgment, action, and avoidance. Each segment of a scenario includes two detailed natural language captions, one for pedestrian behavior and one for vehicle behavior, averaging around 59 words and based on a structured checklist of over 170 items. Furthermore, to support structured reasoning, the dataset includes approximately 180 VQA items and over 3,000 additional annotated videos from BDD100K. Caption evaluation uses LLMScorer, a semantic similarity metric based on large language models, which captures fine-grained alignment between generated and reference descriptions. An example sample can be seen at~\cref{fig:data_overview}.

\subsection{Evaluation Results}

Since our vision encoder is frozen and not being trained during the training process, we decided to extract visual features of videos, save them to files, and use those features in our training process instead of re-running the whole visual encoder every time. By doing this, we preserve more VRAM for training the LLM and the adapter. For the experiment, both the training and testing processes are done with the hardware, including 30GB RAM and a GPU P100 with 16GB VRAM.

Our model was trained on the WTS dataset twice. In the first phase, we trained the model on an internal dataset, which represents a small subset of the full dataset. The training configuration included 20 epochs, a learning rate of 5e-5, the Adam optimizer, a batch size of 2, an evaluation step of 100, and a gradient accumulation step of 2. The results of this training phase are presented in \cref{tab:first_evaluation_metrics}. Although the model was trained on a limited internal dataset and was expected to perform poorly on the external dataset, the results suggest otherwise. The model achieved a high score, nearly 24, on a new dataset with a different context. However, its performance on the external dataset still lags behind the results on the internal dataset. The strong performance after the initial training indicates the robustness of the architecture. Despite being trained on a smaller dataset and for more epochs, the model did not overfit and maintained strong generalization on the external test set. This behavior may be attributed to the event matching component and the diverse information captured during training $\mathcal{L}_M$.

\begin{table}[h]
\centering
\begin{tabular}{lcccc}
\hline
          & \textbf{Internal}  & \textbf{External} \\
\hline
\textbf{BLEU}      & 0.2117    & 0.1654 \\
\textbf{METEOR}    & 0.4281    & 0.3776 \\
\textbf{ROUGE-L}   & 0.4259    & 0.3820 \\
\textbf{CIDEr}     & 0.4559    & 0.3214 \\
\textbf{S}         & 27.7835   & 23.9299 \\
\hline
\textbf{Average S} & \multicolumn{2}{c}{25.8567} \\
\hline
\end{tabular}
\caption{Evaluation metrics for internal and external scores after the first train (20 epochs in 5 hours).}
\label{tab:first_evaluation_metrics}
\end{table}

The same configuration is used for the second training except for the training epoch, which is \textbf{4} in the second training. The primary difference between this training and the first one is that we increased the amount of data, which now includes normal multi-view camera data, normal trimmed data, and external data. The result of this training is presented in \cref{tab:evaluation_metrics}. This result shows the significant improvement of the model in both internal and external test datasets. The result has improved by over \textbf{2 points} for the internal dataset; for the external dataset, the training increases the overall score by more than 5 points. Moreover, after this training, the ability of models to generate descriptions for internal and external data becomes extremely similar. This proved that training models on more data generally leads to better performance.

\begin{table}[h]
\centering
\begin{tabular}{lcccc}
\hline
          & \textbf{Internal}  & \textbf{External}  \\
\hline
\textbf{BLEU}      & 0.2663    & 0.2630 \\
\textbf{METEOR}    & 0.4637    & 0.4684 \\
\textbf{ROUGE-L}   & 0.4587    & 0.4611 \\
\textbf{CIDEr}     & 0.9502    & 1.1400 \\
\textbf{S}         & 32.0934   & 32.6614 \\
\hline
\textbf{Average S} & \multicolumn{2}{c}{32.3774} \\
\hline
\end{tabular}
\caption{Evaluation metrics for internal and external scores of Our Method (4 more epochs in 11 hours).}
\label{tab:evaluation_metrics}
\end{table}

The detail of the weight of each module in the model is shown in \cref{tab:model_weights}. Despite the entire model having nearly 3 billion parameters, the visual encoder is not used during training but for the video pre-processing step, which makes the training process only apply to a much smaller model with about 1.5 billion parameters. Moreover, LoRA layers and the Embedding layer only contain an insignificant number of parameters, and the Adapter has roughly 144 million trainable parameters, which is not very large. The embedding layer in our connector, which is used to get information on the object (pedestrian and vehicle) and segment, is not included in this report due to its insignificant number of parameters.

\begin{table}[h]
\centering
\begin{tabular}{@{}ll@{}}
\toprule
\textbf{Component} & \textbf{Weight} \\
\midrule
Visual encoder     & 1B1 \\
Adapter            & 144M \\
Phi 1.5 with LoRA  & 1B4 \\
\midrule
Total              & 2B7 \\
\bottomrule
\end{tabular}
\caption{Model Component Weights}
\label{tab:model_weights}
\end{table}

\subsection{Methods comparison}

\begin{table*}[ht]
\centering
\label{tab:track2-results}
\begin{tabular}{lcccccc}
\toprule
\textbf{Method} & 
\rotatebox{45}{\textbf{BLEU-4}} & 
\rotatebox{45}{\textbf{METEOR}} & 
\rotatebox{45}{\textbf{ROUGE-L}} & 
\rotatebox{45}{\textbf{CIDEr}} & 
\rotatebox{45}{\textbf{Score}} & 
\rotatebox{45}{\textbf{Model Size}} \\
\midrule
CityLLaVA & 0.278 & 0.477 & 0.470 & 1.130 & 33.43 & $\geq$34B \\
Divide\&Conquer & -- & -- & -- & -- & 32.89 & $\geq$14B \\
PDVC + CLIP & 0.201 & 0.412 & 0.442 & 0.557 & 29.00 & $\geq$2$\times$1B \\
Multi-perspective + Refinement & -- & -- & -- & -- & 22.67 & $\geq$2B \\
\midrule
\textbf{Ours} & 0.2647 & 0.4660 & 0.4599 & 1.0451 & 32.38 & 2.7B \\
\textbf{Ours (w/o Event Matching)} & 0.2603 & 0.4512 & 0.4461 & 0.9056 & 31.20 & 2.7B \\
\bottomrule
\end{tabular}
\caption{Comparison of methods on the WTS dataset. Due to the insufficient information on the two methods, PDVC + CLIP and Multi-perspective + Refinement, we can only estimate their weights. All scores follow the official metric combination: \textit{Score} = \((\text{BLEU-4} + \text{METEOR} + \text{ROUGE-L} + 0.1 \times \text{CIDEr}) / 4 \times 100\).}
\end{table*}

In this section, we compare our method with some existing models on the WTS dataset. As can be seen from~\cref{tab:track2-results}, two models with the best performance used a very large language model to generate descriptions. Meanwhile, methods using smaller LLMs are usually complex, less flexible, and produce worse performance. Our method, despite being unable to overcome large models like CityLLaVA and Devide\&Conquer, our model still has significantly better performance than low-resource models and its performance is on par with methods using very large LLMs. Furthermore, unlike methods like PDVC+CLIP~\cite{shoman2024enhancing}, we do not split our model into two for each scenario, making our method more flexible in general use cases. Lastly, the reported results show that using the Event Matching method increases the model performance by over one point in general, showing its effectiveness and potential application.

\section{Conclusion}

In this paper, we present PerspectiveNet, a model to generate long and detailed descriptions given videos or multiple-viewpoint videos. The model combines a pretrained vision transformer (ViT), a newly constructed connector from two Perceiver modules, and a large language model. Despite its small size and not being pre-trained on any visual information, this model can still achieve a high score on the WTS dataset, indicating its potential in similar tasks.

\section*{Limitation}

Even though this model is small and can generate long fine-grain descriptions on multiple-view cameras and videos, this work has some limitations.

First, this model is constructed to work only with multiple cameras that start at the same time. If any of them start at a different time, the frame feature will not be aligned, which may lead to wrong information comprehension by the model. As a result, the generated descriptions can be unpredictable, and the model should not be used for data like that.

Finally, while bounding boxes can provide more context about pedestrians and vehicles, our solution did not include that important information, which makes the model focus on the wrong target to generate descriptions. This can be more severe if many people and vehicles are on the scene.

\section*{Potential risks}
This work is LLM-based, which may result in some wrong information, like generating wrong or biased descriptions.

\bibliography{custom}

\appendix

\end{document}